\documentclass[runningheads]{llncs}
\usepackage[T1]{fontenc}
\usepackage{graphicx}
\usepackage{amsfonts}
\usepackage{amsmath}
\usepackage{makecell} 
\usepackage{multirow} 
\usepackage{graphicx} 
\usepackage{booktabs} 
\usepackage{siunitx}  
\usepackage[table]{xcolor} 
\usepackage{hyperref}

\usepackage{tikz}
\usetikzlibrary{shapes.geometric, arrows.meta, positioning, shadows}

\newcommand{\boldx}{\mbox{${\mathbf x}$}}

\begin{document}
\title{Assessing an evolutionary search engine for small language models, prompts, and evaluation metrics}
\titlerunning{Assessing an evol. search engine for SLM's, prompts, and eval. metrics}
%
\author{Cláudio Lúcio do Val Lopes\inst{1}\orcidID{0000-0003-1655-2283} 
\and \\ 
Lucca Machado da Silva\inst{1}\orcidID{0009-0002-7191-6756}
}

\authorrunning{Cláudio L V Lopes and L. Machado da Silva}
%
\institute{A3Data  - \url{http://www.a3data.com.br}\\
Correspondence to: \email{claudio.lucio@a3data.com.br}\\
 }
\maketitle              
\begin{abstract}
 The concurrent optimization of language models and instructional prompts presents a significant challenge for deploying efficient and effective AI systems, particularly when balancing performance against computational costs like token usage. This paper introduces and assesses a bi-objective evolutionary search engine designed to navigate this complex space, focusing specifically on Small Language Models (SLMs). We employ the NSGA-II algorithm and prompt grammar to simultaneously optimize for task accuracy and token efficiency across some reasoning tasks. Our results successfully identify diverse, high-performing model-prompt combinations, quantitatively revealing the critical trade-off between the two objectives. This research highlights task-specific affinities between particular SLMs and prompt structures (e.g., instructions, context, chain of thought). The generated practical Pareto fronts offer decision-makers a portfolio of optimized solutions adaptable to their specific constraints. This automated approach moves beyond traditional manual tuning, providing a foundational framework for discovering effective human-AI interaction patterns.

\end{abstract}
\section{Introduction}

In recent years, Large Language Models (LLMs) and Small Language Models (SLMs) have gained increasing relevance across numerous applications. The apparent ability of these models to solve tasks that require intricate linguistic comprehension has driven their widespread adoption. However, the quality of responses generated by LLMs/SLMs is profoundly influenced by the way they are queried~\cite{10.1145/3638529.3654049}, a process known as prompt engineering.

Manual prompt engineering is often tedious and requires considerable human effort and expertise~\cite{wang2024promptagent}. Furthermore, optimal prompts are not universally transferable across different models or tasks, necessitating repeated design efforts. This challenge is particularly pronounced for Small Language Models (SLMs)~\cite{10897262}, which, by their nature, are more sensitive to the nuances of prompt engineering than their larger counterparts, making prompt optimization even more critical to achieving acceptable performance. The search for an effective prompt is inherently difficult due to the complex internal workings of the models and the vast search space of possible natural language formulations~\cite{Automatic_prom_Grad,NEURIPS2024_46ed5038}.

Quality indicators or metric evaluations help guide the navigation process in the search space. However, considering the generative nature of language models, there is often no single `correct' output for a given input, particularly for open-ended tasks like text summarization or creative writing~\cite{10.1145/3641289}. Evaluating the quality of generated text requires checking aspects like fluency, coherence, relevance, and creativity, which are inherently difficult to quantify. In response to these complexities, the field is observably shifting away from evaluations based on single metrics towards a more multi-dimensional approach~\cite{liang2023holisticevaluationlanguagemodels} that explicitly advocates multi-metric assessment, validating models across dimensions such as accuracy, robustness, fairness, bias~\cite{10.1162/coli_a_00524}, toxicity, and efficiency.

Given these difficulties, automation of the prompt design process emerges as a possible solution. The search for an effective solution can be framed as an optimization problem~\cite{app15031430,cui2025automaticpromptoptimizationheuristic}: to find a prompt combined with a model, a SLM, that maximizes/minimizes some evaluation metrics, obtaining the desired response to the user input. 


This paper addresses the joint challenge of selecting language models and designing optimal prompts. We propose an automated approach using a multi-objective evolutionary optimization algorithm, NSGA-II~\cite{debieee}, to mitigate these difficulties. Given a task, NSGA-II searches for solutions, where each solution is an individual represented by a combination of a prompt generated using a grammar~\cite{10.1145/3638529.3654049,Lourenço2018}), and an SLM. The search concludes with a Pareto set of solutions that achieve the best compromises across multiple evaluation metrics.

Leveraging our evolutionary search, we successfully identified diverse, high-performing prompt-model combinations across various reasoning tasks from the Beyond the Imitation Game benchmark (BIG-bench)~\cite{Srivastava2023BeyondTI}. These findings offer practitioners and decision makers a practical basis for informed decision-making regarding effective SLM deployment.

The paper is structured as follows. Section \ref{Background} offers some background about optimization problems with SLM, prompts, and metric evaluations. Multi- and many-objective evolutionary algorithms are presented, and the problem formulation is also detailed. Section \ref{Proposed approach} describes our novel approach using the NSGA-II algorithm to automatically search for optimal SLM and prompt combinations based on evaluation criteria. Section \ref{Experiments} details the empirical setup, including the BIG-bench datasets and the experimental procedures. Section \ref{Results and discussions} analyzes and interprets the outcomes of our experiments. Finally, Section \ref{Conclusion and future works} summarizes our findings and suggests areas for future research.

\section{Background}  \label{Background}
\subsection{Language models and prompts as an optimization problem}

Starting with prompt optimization, it encompasses a range of strategies, from manual crafting by human experts to fully automated procedures. Historically, interacting with LMs relied heavily on manual prompt engineering. This involves a human expert iteratively designing, testing, and refining input phrases or instructions to guide the model's behavior. Some common manual techniques include: zero-shot~\cite{lepagnol-etal-2024-small}, few-shot, In-Context Learning (ICL), Chain-of-Thought (CoT), instruction tuning prompts, and self-consistency (details in~\cite{Ferraris_Audrito_Caro_Poncibò_2025}). The manual nature of these techniques makes achieving optimal performance a significant undertaking. 

Automated Prompt Optimization (APO) tries to mitigate the limitations of manual engineering. These techniques aim to discover or refine prompts algorithmically, reducing human effort and potentially finding non-intuitive solutions. APO approaches can be broadly categorized: Gradient-based Method, such as in~\cite{Automatic_prom_Grad}, Discrete Search/Optimization Methods, as in~\cite{chen2025}, and LLM-as-judge~\cite {wang2024promptagent}.

Evolutionary algorithm techniques are classified as search and optimization methods. Algorithms such as particle swarm optimization and genetic algorithms are employed to evolve populations of prompts towards enhanced performance. Notable examples include Evoprompt~\cite{10.1145/3638529.3654049} and InstOptima~\cite{YangL23instoptima}.

EvoPrompt utilizes a grammar-based evolutionary method to optimize prompts for Large Language Models (LLMs). This technique automatically evolves the structure of prompts to enhance the performance of an LLM on a given task. The core of EvoPrompt's methodology is the use of a predefined grammar that dictates the possible structures of a prompt, which are then evolved over generations. These evolutionarily optimized prompts have been shown to yield better results than traditional prompting methods, considering some models and tasks~\cite{10.1145/3638529.3654049}. EvoPrompt also examines the effectiveness of different prompt components for particular model-task pairings. However, a significant limitation of this approach is its lack of an automatic search for different models; the optimization is performed for a predetermined model.

InstOptima presents a different strategy by framing instruction generation as a multi-objective optimization problem. This method simultaneously targets multiple objectives, including performance, length, and perplexity. A key feature of InstOptima is its use of LLMs as evolutionary operators to generate and refine instructions. While this allows for a dynamic and powerful optimization process, it also introduces considerable drawbacks. The reliance on LLMs for evolutionary operations leads to significant token costs and can create complexities in managing the evolutionary dynamics. Similar to EvoPrompt, InstOptima's optimization process does not include the selection of the language model itself.

Our method adopts a grammar-based approach, similar to Evoprompt, which circumvents the cost associated with using LLMs as evolutionary operators. Furthermore, in contrast to InstOptima, we propose incorporating language models as part of the evolutionary process. Indeed, we will demonstrate that the language model forms an integral part of the individual within the evolutionary process.

\subsection{Multi- and Many-Objective Evolutionary algorithms}
Many real-world optimization problems are made up of multiple conflicting objectives. Although traditional approaches can combine objectives into one single solution and solve the problem, several multi- and many-objective optimization methods have proven to be efficient techniques to deal with the true multi-objective nature of such problems. 

In general, a multi- or many-objective optimization problem (MOP, MaOP, respectively) includes $N$ decision variables, \boldx, from a feasible decision space $\Omega \subseteq \mathbb{R}^{N}$, and a set of $M$ objective functions. Without loss of generality, the minimization of an MOP can be simply defined as:

\begin{align}\label{MOP_definition}
\mbox{Minimize} \quad F(\boldx) = {[f_{1}(\boldx), \ldots, f_{M}(\boldx)]}^{T}, \quad  \boldx \in \Omega.
\end{align}

$F: \Omega \rightarrow \Theta \subseteq \mathbb{R}^{M}$ is a mapping from the feasible decision space $\Omega$ to vectors in the $M$-dimensional objective space $\Theta$. 
When $M \geq 4$, the problem is commonly called a many-objective problem.

There are different ways to deal with multi- and many-objective optimization problems. Evolutionary optimization is one of the methods available and used to treat multi- and many-objective problems (EMO, and EMaO, respectively). They have shown certain success in finding well-converged and well-diversified non-dominated solutions~\cite{deb-book-01}.

Evolutionary or population-based optimization methods collect information from the objective functions at more than one point in the solution space; each one is called an individual. The individuals evolve towards the next generation through the information collected, using evolutionary procedures that mimic evolution concepts in nature, such as selection, mutation, and crossover.

Some successful algorithms using this inspired evolutionary approach include, for example, the non-dominated sorting genetic algorithm: NSGA-II~\cite{debieee} and NSGA-III~\cite{nsga3-1}; and some advanced evolutionary frameworks to deal with hard problems~\cite{debClaudio}; and many others.

\subsection{Problem formulation}

Consider a set of $N$ different LLM or SLM models $\mathcal{M} = \{M_1, ..., M_N\}$. Each model $M_i: Q \times P \rightarrow A$ can be abstracted as a function that maps a query $q \in Q$ and a prompt $p \in P$ to an answer $a \in A$.

Now consider that each answer must be evaluated by a quality indicator or evaluation metric $I: Q \times P \times \mathcal{M} \rightarrow \mathbb{R}$, it takes a query $q \in Q$, a prompt $p \in P$, and $m \in \mathcal{M}$ and assess the answer that generates a real value;

\begin{definition}
Quality indicator: An $I$ is a function $I:  Q \times P \times \mathcal{M} \rightarrow \mathbb{R}$, which assigns to each vector of tuples $[(q_1, p_1, m_1), (q_2, p_2, m_2), \dots, (q_K, p_K, m_K)]$ a real value $I([(q_1, p_1, m_1), (q_2, p_2, m_2), \dots, (q_K, p_K, m_K)])$.
\end{definition}

Note that the literature employs a diverse range of quality indicators for model evaluation. Consequently, we may encounter $L$ distinct quality indicators, so for a tuple of $(q, p, m)$, it is possible to have a vector of quality indicators such as $\Theta = [I_1, I_2, \dots, I_L]$. 

Our goal is to find the optimal combination of a prompt $p$ and a model $m$ that simultaneously minimizes all $L$ quality indicators when evaluated over the query set $Q$.

The decision variable for our optimization problem is the pair $\boldx = (p, m)$, representing the choice of a specific prompt $p$ from the prompt space $P$ and a specific model $m$ from the model set $\mathcal{M}$.

The feasible decision space $\Omega$ is the Cartesian product of the prompt space and the model set:
$$ \Omega = P \times \mathcal{M} $$

For a given decision variable $\boldx = (p, m)$, we define the vector of query-prompt-model tuples used for evaluation as:
$$ \mathcal{T}(p, m, Q) = [(q_1, p, m), (q_2, p, m), \dots, (q_K, p, m)] $$
Note that the prompt $p$ and model $m$ are fixed for all queries $q_k$ in this evaluation vector.

The $L$ objective functions, $f_1, \dots, f_L$, correspond directly to the $L$ quality indicators evaluated on the results generated for the query set $Q$ using the chosen prompt $p$ and model $m$. Therefore, the $l$-th objective function is:
$$ f_l(\boldx) = f_l(p, m) = I_l(\mathcal{T}(p, m, Q)), $$
where $l = 1, \dots, L$.

The multi-objective optimization problem is then formulated as the minimization of the vector of these quality indicators:

\begin{align}\label{PromptMOP_definition}
\mbox{Minimize} \quad F(p, m) = & {[f_{1}(p, m), \ldots, f_{L}(p, m)]}^{T} \\
 = & {[I_1(\mathcal{T}(p, m, Q)), \ldots, I_L(\mathcal{T}(p, m, Q))]}^{T}, \\
\mbox{subject to} \quad & (p, m) \in P \times \mathcal{M}.
\end{align}

This formulation seeks to find the set of non-dominated solutions (Pareto front) in terms of prompt-model pairs $(p, m)$, where each solution represents a trade-off among the $L$ different quality indicators being minimized. Evolutionary algorithms like NSGA-II are well-suited to finding an approximation of this Pareto front.

\section{Proposed approach} \label{Proposed approach}
\subsection{General solution}

\begin{figure} [!htb]
\centering
\includegraphics[scale=0.45]{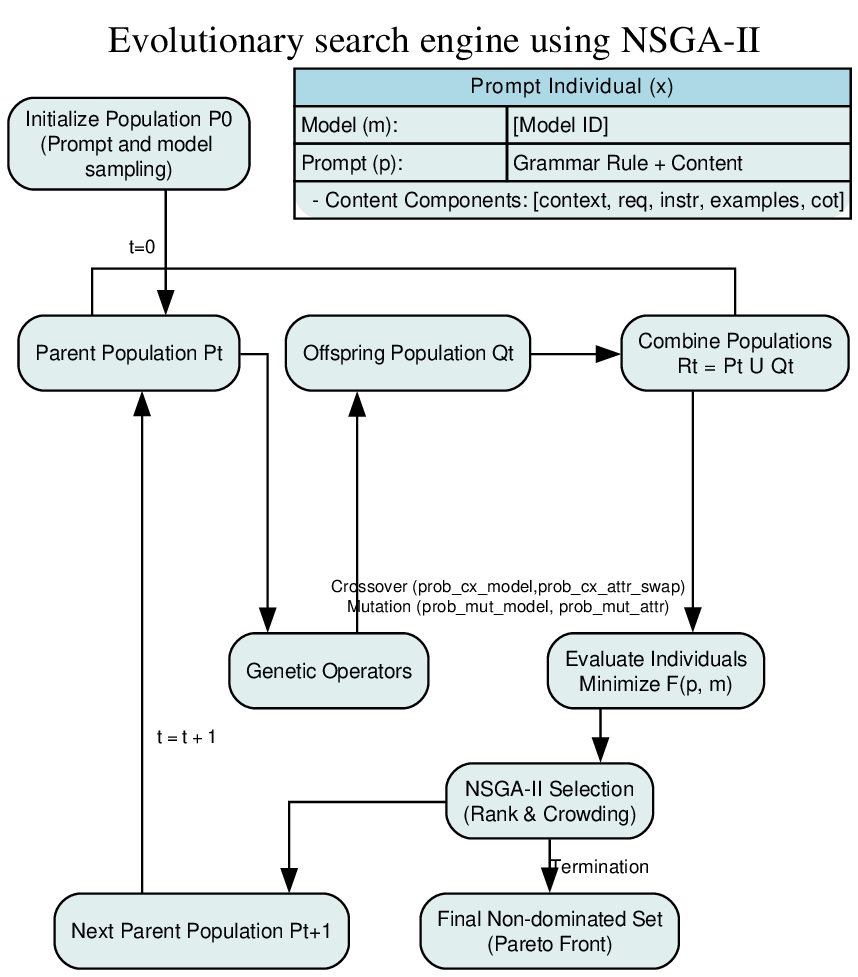}
\caption{Overview of the multi-objective evolutionary prompt optimization process using NSGA-II. (1) An initial population ($P_0$) of \textit{Prompt Individuals} generated via sampling, combining a language model ID and a prompt structure (grammar rule and content). (2) The Parent Population ($P_t$) enters the evolutionary loop. (3) Genetic operators (Crossover and Mutation) are applied based on their probabilities. These operators modify the model choice and/or prompt components using specific probabilities ($p_{cx\_model}$, $p_{cx\_attr}$, $p_{mut\_model}$, $p_{mut\_param}$). (4) The combined population ($R_t = P_t \cup Q_t$) is evaluated by running the prompts against a dataset sample to obtain objective values $F(p, m)$, e.g., minimizing $1 - \text{Accuracy}$ and average tokens. (5) NSGA-II selects the individuals for the next Parent Population ($P_{t+1}$). (6) The cycle repeats for a predefined number of generations ($t=t+1$). (7) Upon termination, the final non-dominated set, representing the best trade-off solutions found (Pareto front), is output.}
\label{fig:prompt_evolution_detailed}
\end{figure}

To automatically discover effective prompts for interacting with different language models, we employed an evolutionary optimization approach. Figure \ref{fig:prompt_evolution_detailed} illustrates this process, which uses, for example, the NSGA-II algorithm.

The central concept involves treating distinct combinations of language models and prompt structures as `individuals' within a population. Each individual represents a unique approach to instructing a language model to perform a specific task using a prompt. 

The process begins by creating an initial diverse population (P0) of these prompt-model individuals randomly. This population then enters an iterative evolutionary cycle. In each cycle (generation), the current set of individuals (Parent Population Pt) is used to create new potential solutions (Offspring Population Qt) through simulated crossover and mutation. Crossover might mix elements from two successful parent prompts, while mutation might slightly alter a prompt's grammar components or even change the language model being used. 

The newly created offspring are combined with the parents. All individuals in this combined group are then evaluated using each prompt-model combination to generate answers for a set of representative task examples (a dataset sample). We then calculate the quality indicators, such as the accuracy, and the number of tokens processed. Our goal is to find prompts that are both highly accurate and efficient, for example.

Based on this evaluation, the NSGA-II selection process chooses the fittest individuals to form the next generation's Parent Population (Pt+1). This selection favors individuals that perform well on both accuracy and efficiency, particularly those that represent the best trade-offs.

This cycle of generating offspring, evaluating, and selecting repeats for a set number of generations, loops back. When the process finishes, the final output is the set of non-dominated solutions – the best trade-offs found, known as the Pareto front.

A prompt itself is formally defined by two components: a grammatical rule that dictates the inclusion and ordering of elements like instructions, examples, or context, and the specific textual content used for each of these defined parts.

We use grammars as proposed in~\cite{10.1145/3638529.3654049} and 
in~\cite{Lourenço2018}. A Backus–Naur Form (BNF) grammar is designed to effectively navigate the vast prompt space for Language Models in a sample. Recognizing that searching the entire space of potential English sentences is computationally impractical, the grammar provides a structured approach to guide the prompt search in the evolutionary process. It functions as a Context-Free Grammar (CFG)~\cite{Lourenço2018}, defining how prompts can be constructed as compositions of various parts, drawing inspiration from existing prompt engineering literature.

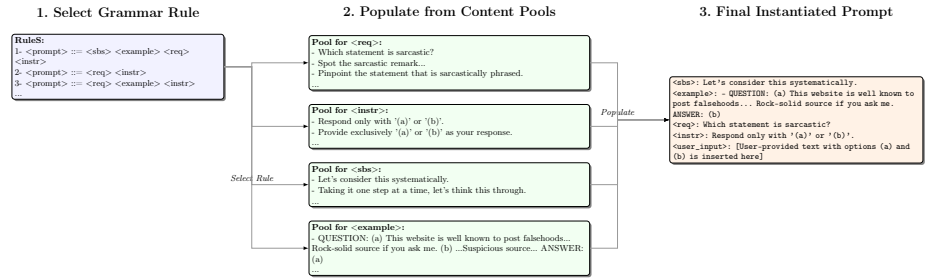
\begin{figure}
\resizebox{\textwidth}{!}{

\begin{tikzpicture}[
    node distance=1.5cm and 1cm,
    scale=1,
    stage/.style={
        font=\Large\bfseries,
        text=black!60!black
    },
    rule_box/.style={
        rectangle,
        draw,
        thick,
        rounded corners=3pt,
        fill=blue!5,
        text width=7.5cm,
        align=left,
        minimum height=2cm,
        drop shadow={opacity=0.3}
    },
    pool_box/.style={
        rectangle,
        draw,
        thick,
        rounded corners=3pt,
        fill=green!5,
        text width=10cm,
        align=left,
        minimum height=1.5cm,
        drop shadow={opacity=0.3}
    },
    final_prompt/.style={
        rectangle,
        draw,
        thick,
        rounded corners=3pt,
        fill=orange!10,
        text width=9cm,
        align=left,
        minimum height=3cm,
        font=\ttfamily,
        drop shadow={opacity=0.3}
    },
    arrow/.style={
        -{latex[length=3mm]},
        thick,
        draw=gray!80
    }
]

\node (stage1) [stage] {1. Select Grammar Rule};
\node (rule2) [rule_box, below=0.5cm of stage1] {
    \textbf{RuleS:} \\
    1- $<$prompt$>$ ::= $<$sbs$>$ $<$example$>$ $<$req$>$ $<$instr$>$\\
    2- $<$prompt$>$ ::= $<$req$>$ $<$instr$>$\\
    3- $<$prompt$>$ ::= $<$req$>$ $<$example$>$ $<$instr$>$\\
    ...
};

\node (stage2) [stage, right=5cm of stage1] {2. Populate from Content Pools};

\node (req_pool) [pool_box, below=0.5cm of stage2] {
    \textbf{Pool for $<$req$>$:}  \\
    - Which statement is sarcastic? \\
    - Spot the sarcastic remark... \\
    - Pinpoint the statement that is sarcastically phrased.\\
    ...
};
\node (instr_pool) [pool_box, below=0.5cm of req_pool] {
    \textbf{Pool for $<$instr$>$:}  \\
    - Respond only with '(a)' or '(b)'. \\
    - Provide exclusively '(a)' or '(b)' as your response.\\
    ...
};

\node (sbs_pool) [pool_box, below=0.5cm of instr_pool] {
    \textbf{Pool for $<$sbs$>$:}  \\
    - Let's consider this systematically. \\
    - Taking it one step at a time, let's think this through. \\
    ...
};
\node (example_pool) [pool_box, below=0.5cm of sbs_pool] {
    \textbf{Pool for $<$example$>$:}\\
    - QUESTION: (a) This website is well known to post falsehoods... Rock-solid source if you ask me. (b) ...Suspicious source... ANSWER: (a) \\
    ...
};

\node (stage3) [stage, right=5cm of stage2] {3. Final Instantiated Prompt};
\node (final) [final_prompt, below=2cm of stage3] {
    \textbf{<sbs>}: Let's consider this systematically.
    
    \textbf{<example>}: - QUESTION: (a) This website is well known to post falsehoods... Rock-solid source if you ask me. \\ANSWER: (b)
    
    \textbf{<req>}:  Which statement is sarcastic? \\
    
    \textbf{<instr>}: Respond only with '(a)' or '(b)'.
    
    \textbf{<user\_input>}: [User-provided text with options (a) and (b) is inserted here]
};


\draw [arrow] (rule2.east) -- ++(1,0) |- node[above, midway, font=\itshape] {Select Rule } (sbs_pool.west);
\draw [arrow] (rule2.east) -- ++(1,0) |- (example_pool.west);
\draw [arrow] (rule2.east) -- ++(1,0) |- (req_pool.west);
\draw [arrow] (rule2.east) -- ++(1,0) |- (instr_pool.west);

\draw [arrow] (sbs_pool.east) -- ++(1,0) |- node[above, midway, font=\itshape] {Populate} (final.west);
\draw [arrow] (example_pool.east) -- ++(1,0) |- (final.west);
\draw [arrow] (req_pool.east) -- ++(1,0) |- (final.west);
\draw [arrow] (instr_pool.east) -- ++(1,0) |- (final.west);

\end{tikzpicture}
}
\caption{A schematic of the grammar-based prompt generation process. The process begins with \textbf{(1)} the selection of a structural grammar rule composed of non-terminal symbols (e.g., \textit{<sbs>}, \textit{<req>}). \textbf{(2)} This rule dictates which content pools of terminal symbols (actual text strings) are used. \textbf{(3)} Finally, one variant from each required pool is selected to assemble, or instantiate, the complete, ready-to-use prompt.}
\label{fig:grammar}
\end{figure}

The grammar guides prompt construction through a three-stage process, as illustrated in Figure \ref{fig:grammar}. The process is initiated in \textbf{1.} with the selection of a grammar rule, such as a basic prompt (\textit{<zero>}) or one including examples (\textit{<example>}). These rules define the prompt's high-level structure by combining various non-terminal symbols that represent specific textual elements. These elements include generic components like \textit{<context>} and \textit{<sbs>} (`think step by step'), as well as dataset-specific components like the base request (\textit{<req>}) and examples (\textit{<example>}). As shown in \textbf{2.}, each non-terminal symbol in the chosen rule is then linked to a dedicated ´content pool' of terminal symbols, which are the actual texts  used to populate the prompt. According to the design, all non-terminal grammar elements, except for examples, are associated with up to 100 semantically similar variants previously generated by an LLM. For the \textit{<example>} component, the content is drawn directly from the dataset. The process culminates in \textbf{3.}, where a single terminal string is selected from each required pool to construct the final, instantiated prompt, which is then ready to be combined with the user's input.

This evolutionary process search for prompt structures and SLM to provide a range of high-performing solutions, allowing a decision maker to choose the one that best fits their specific needs regarding the balance between accuracy and computational cost.

\subsection{Evolutionary aspects}
We employ two standard genetic operators, mutation and crossover, to explore the search space of different prompt structures, content variations, and language models.

The mutation operator introduces random changes to an individual, which represents a specific prompt-model pair. There is a mutation probability for each. The model mutation probability is associated with the language model. 
When a model mutation occurs, a different model is randomly selected from the predefined pool of available models. This allows the search to explore the effectiveness of the same prompt structure with different underlying language models. Prompt mutation probability is attached to the content of one of the prompt's grammatical components (such as \textit{<context>}, \textit{<req>}, \textit{<instr>}, \textit{<examples>}, or \textit{<cot>}). Suppose a prompt mutation occurs for a specific component. In that case, the existing text is replaced with a different, randomly selected text option of the same component type, drawn from the pool of options generated. 

The crossover operator combines genetic material from two parent individuals to create two new offspring individuals. This operator is applied to selected pairs of parents, with two distinct probabilities. With model crossover probability, the language model associated with the two parent individuals is potentially swapped between the resulting offspring: offspring 1 receives parent 2's model, and offspring 2 receives parent 1's model. Otherwise, each offspring inherits its primary parent's model.

Prompt component crossover probability exchanges textual content for prompt components between the offspring, but with a crucial condition: a swap for a specific component only occurs if both parents possess non-empty content for that component. This respects the potentially different grammatical structures of the parents. For example, if both parents have a <context> part, the context text is swapped between the offspring. If one parent's rule includes <context> but the other's does not, the context component is not swapped. This conditional uniform crossover allows beneficial textual elements to be shared between individuals, even if their overall prompt structures differ slightly. The prompt structural component and user input are never exchanged during crossover.

These operators, working together within the NSGA-II framework, enable a dynamic exploration of both the language model and the diverse prompt contents, searching for combinations that yield high-quality indicators.

\section{Experiments} \label{Experiments}
\subsection{Datasets}
Following the experiments proposed in~\cite{10.1145/3638529.3654049}, six specific reasoning tasks from BIG-bench Lite~\cite{Srivastava2023BeyondTI} were used for our experiments. A key characteristic of these chosen tasks is that they all require a binary classification output (e.g., `yes'/`no', `true'/`false', `(a)'/`(b)', `correct'/`incorrect'). This binary nature allows for straightforward accuracy calculation and establishes a random baseline performance of 0.50 accuracy.
The specific tasks with their descriptions are as follows:


\begin{itemize}
    \item Causal Judgment: Requires identifying causal relationships versus mere correlations. Contains 180 instances;
    \item Hyperbaton: Tests understanding of natural English word order by choosing the more standard sentence between two options involving adjective ordering. Contains approximately 50,000 instances;
    \item Implicatures: Focuses on pragmatic understanding, predicting whether a response implies a `yes' or `no'. Contains 483 instances;
    \item Logical Fallacy Detection: Requires identifying whether an argument contains a formal or informal logical fallacy. Contains 2790 instances;
    \item Navigate: Assesses spatial reasoning by determining if following a set of navigation instructions returns the agent to the starting point. 990 instances;
    \item Winowhy: Evaluates the model's reasoning behind resolving Winograd Schemas, which involve pronoun disambiguation. Contains 2855 instances;
\end{itemize}

\subsection{Methodology}

We address a multi-objective optimization problem with two conflicting objectives for simultaneous minimization: I) Inverse Accuracy, calculated as (1 - Accuracy), where accuracy is determined by comparing the model's generated answer (e.g., `yes', `no', `(a)', `(b)') against the ground truth for each task instance in the dataset; and II) Average Total Tokens, representing the average number of tokens processed per task instance during evaluation. This includes tokens in the input prompt and the model's generated output. 


To demonstrate the effectiveness of the evolutionary search even with less powerful models, we focused exclusively on Small Language Models (SLMs) with parameter counts generally under 2 billion. The specific models included in the evolutionary search space were: DeepSeek-R1-Distill-Qwen-1.5B~\cite{deepseekai2025deepseekr1incentivizingreasoningcapability}, Qwen2.5-1.5B-Instruct~\cite{qwen2.5}, Llama-3.2-1B-Instruct ~\cite{grattafiori2024llama3herdmodels}, Nvidia OpenMath-Nemotron-1.5B~\cite{moshkov2025aimo2}, and Google gemma-3-1b-it~\cite{gemma_2025}.
The models were chosen for their readily available implementations on HuggingFace, a factor that significantly contributes to the reproducibility of this work.



The search space for prompts was defined by a predefined BNF grammar, specific to each task type. As previously explained, the textual options for components like \textit{<context>}, \textit{<req>}, \textit{<instr>}, etc., within these grammars were initially generated using an auxiliary LLM to provide diverse and relevant content possibilities. The evolutionary algorithm explores combinations of grammar rules and these content options.

We employed the NSGA-II algorithm, specifically using the implementation provided by the pymoo Python library \cite{pymoo}. The population evolved over 10 generations, with each generation maintaining 30 distinct prompt-model individuals. During the crossover operation between two parent individuals, there was a 0.9 probability of swapping their associated language models. A similar probability of 0.9 was applied to exchange the content of each shared, non-empty prompt component (such as context or instruction) between the offspring. For the mutation operation, the probability of either a model change or a modification to the content of a single prompt parameter was set to 0.2.

Each individual's fitness (accuracy and token count) was estimated based on its performance on a randomly selected sample of 100 instances from the full dataset for that task. The same random seed was used for sampling within each independent run to ensure a fair comparison between individuals in that run.

The entire evolutionary process (10 generations, 30 individuals) was repeated 11 independent times for each task, starting from different random initial populations and using different random seeds for the evolutionary operators and evaluation sampling between runs. This helps ensure the robustness and consistency of the results.

After completing all 11 independent runs for a given task, the non-dominated solutions found at the end of each run were collected. This combined set of solutions was then deduplicated, removing individuals with identical models and prompt structures (ignoring the input placeholder). Finally, a non-dominated sorting process was applied to this unique, aggregated set to identify the final global Pareto front, representing the best trade-offs between accuracy and token efficiency discovered across all runs. 

The experiments were done in one Nvidia L40 with 48 GB of GDDR6 memory. The average duration of each experiment per dataset was 489.79 seconds. The code can be accessed \href{https://github.com/ClaudioLucioLopes/EvoMultSearch}{here}.

\section{Results and discussions} \label{Results and discussions}

\begin{figure}[!htb]
\centering
\includegraphics[scale=0.296]{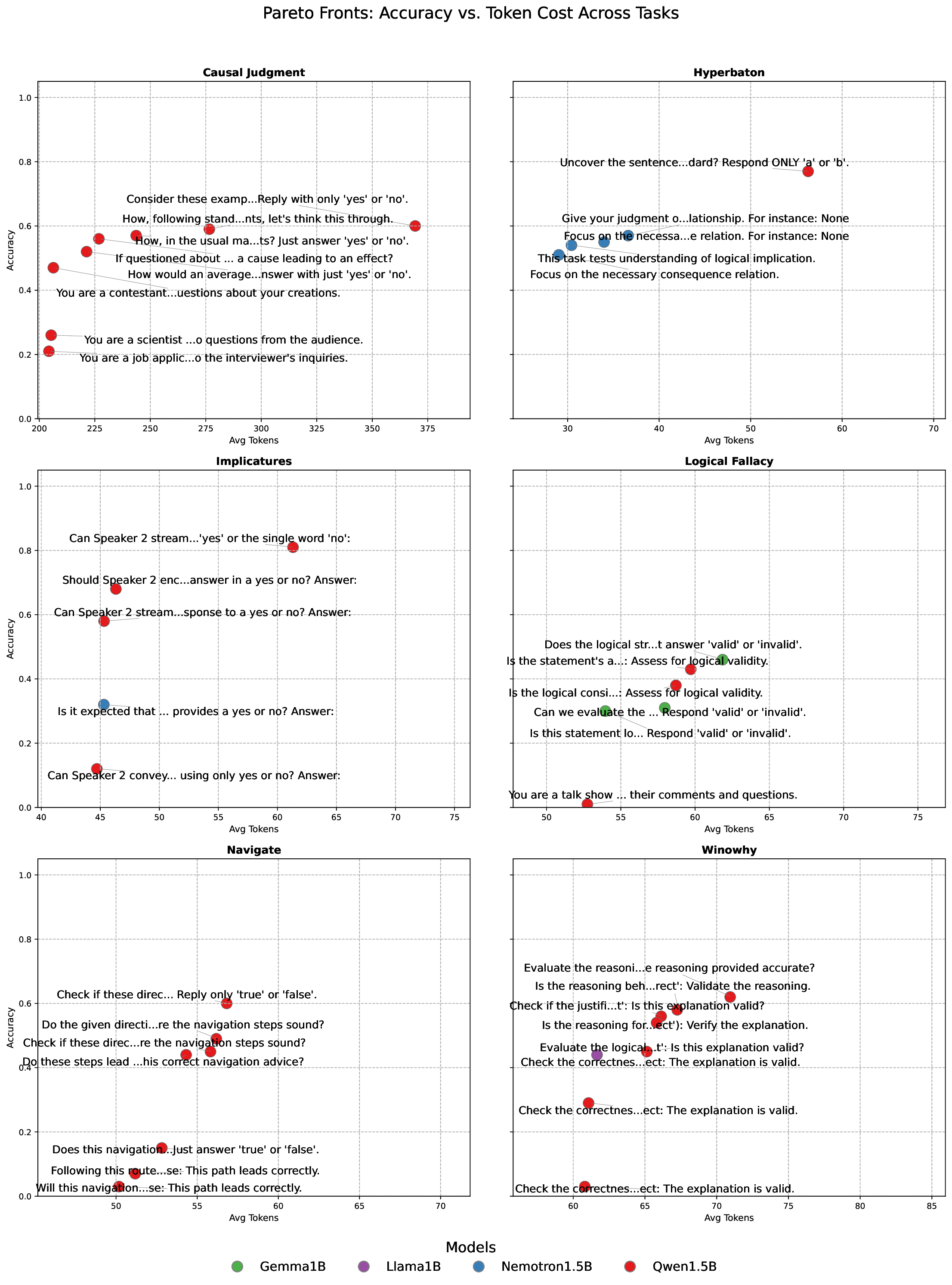}
\caption{Final Pareto fronts obtained from the evolutionary optimization across six different BIG-bench Lite reasoning tasks. Each subplot displays the non-dominated solutions found after aggregating 11 independent runs. The Y-axis represents the average accuracy achieved, while the X-axis the average total token count (input + output) per instance. Points are color-coded by the language model used, as indicated in the shared legend. This visualization highlights the accuracy vs. token cost trade-offs and the performance of different models on each task. The label for each point is the prompt's beginning and ending parts.}
\label{fig:final_pareto_grid}
\end{figure}

After aggregating results from all runs and performing a final non-dominated sort, we obtained the Pareto fronts representing the best trade-offs between maximizing accuracy and minimizing average token count. Figure \ref{fig:final_pareto_grid} visually presents these final Pareto fronts for each task in a 3x2 grid with different models indicated by color. 

Figure \ref{fig:final_pareto_grid} consistently confirms the expected trade-off between accuracy and token efficiency across all six reasoning tasks. Solutions pushing towards higher accuracy (top-left region of each plot) invariably demand a higher average token count, often involving more complex prompt structures or specific models. Conversely, solutions optimized for minimal token usage (bottom-right region) frequently sacrifice significant accuracy (Logical Fallacy, Navigate, Winowhy). This visualization underscores that there is rarely a single `best' model and prompt; instead, the evolutionary search identifies a set of optimal compromises. For example, in Hyperbaton, the highest accuracy solution (0.77) uses 56 tokens, while a much cheaper solution using only 29 tokens still achieves 0.51 accuracy. The choice between these, depends on the user's specific requirements.

Table \ref{tab:pareto_summary_detailed_restructured} complements this by summarizing key characteristics of these fronts, including the dominant models and prompt component, as well as highlighting the specific solutions achieving the highest accuracy, lowest token cost, and a balanced `knee' point for each task.

Visualizing the distribution of the model on the Pareto fronts (Figure \ref{fig:final_pareto_grid}, color-coded points) and summarizing their counts in Table \ref{tab:pareto_summary_detailed_restructured} reveal the strengths of the task-specific model among the tested SLMs: Qwen1.5B demonstrated remarkable performance, dominating the Pareto fronts entirely for Causal Judgment and Navigate, and constituting most solutions for Implicatures and Winowhy. It consistently achieved the highest accuracy points on these tasks. Nemotron1.5B showed a strong presence on the Hyperbaton front, particularly in the lower-token, mid-accuracy range, including the lowest-token and knee solutions. Gemma1B was notably effective for Logical Fallacy Detection. Llama1B appeared less frequently, and the deepseek-ai model did not feature on the final aggregated Pareto fronts for these tasks, suggesting it was generally outperformed in this context.

\begin{table*}[!h]
\centering
\caption{Summary of Final Pareto Front Characteristics for Each BIG-bench Lite Task. `Dominant Components' lists components present in the most frequent prompt grammars on the front, with counts reflecting their frequency. Solution details for the highest accuracy, lowest token cost, and a balanced 'knee' point (a tradeoff) are provided under the `Solution Types' header. The best accuracy and lowest token values are highlighted in bold.}
\label{tab:pareto_summary_detailed_restructured}
\resizebox{\textwidth}{!}{
\begin{tabular}{@{}l|l|l|ccc|ccc|ccc@{}}
\toprule
\multirow{3}{*}{\textbf{Dataset}} & \multirow{3}{*}{\textbf{\makecell[l]{Dominant\\ Models\\(Count)}}} & \multirow{3}{*}{\textbf{\makecell[l]{Dominant\\ Components\\(Count)}}} & \multicolumn{9}{c}{\textbf{Solution Types}} \\ \cmidrule(lr){4-12} 
 &  &  & \multicolumn{3}{c|}{\textbf{Best Accuracy}} & \multicolumn{3}{c|}{\textbf{Lowest Tokens}} & \multicolumn{3}{c}{\textbf{'Knee'}} \\ \cmidrule(lr){4-6} \cmidrule(lr){7-9} \cmidrule(lr){10-12} 
 &  &  & \textbf{Model} & \textbf{Acc.} & \textbf{Tokens} & \textbf{Model} & \textbf{Acc.} & \textbf{Tokens} & \textbf{Model} & \textbf{Acc.} & \textbf{Tokens} \\ \midrule \midrule

Causal Judgm. & \makecell[l]{Qwen1.5B (8)} & \makecell[l]{req (4) \\ instr (4) \\ context (3) \\ cot (2) \\ examples (1)} & Qwen1.5B & \bfseries 0.60 & 369 & Qwen1.5B & 0.21 & \bfseries 204 & Qwen1.5B & 0.56 & 227 \\ \midrule

Hyperbaton & \makecell[l]{Nemotron1.5B (4) \\ Qwen1.5B (1)} & \makecell[l]{context (4) \\ examples (2) \\ cot (2) \\ req (1) \\ instr (1)} & Qwen1.5B & \bfseries 0.77 & 56 & Nemotron1.5B & 0.51 & \bfseries 29 & Nemotron1.5B & 0.54 & 30 \\ \midrule

Implicatures & \makecell[l]{Qwen1.5B (4) \\ Nemotron1.5B (1)} & \makecell[l]{req (5) \\ instr (5)} & Qwen1.5B & \bfseries 0.81 & 61 & Qwen1.5B & 0.12 & \bfseries 45 & Qwen1.5B & 0.68 & 46 \\ \midrule

Logical Fallacy & \makecell[l]{Gemma3-1B (3) \\ Qwen1.5B (3)} & \makecell[l]{req (5) \\ instr (5) \\ context (1)} & Gemma3-1B & \bfseries 0.46 & 62 & Qwen1.5B & 0.01 & \bfseries 53 & Gemma3-1B & 0.30 & 54 \\ \midrule

Navigate & \makecell[l]{Qwen1.5B (7)} & \makecell[l]{req (7) \\ instr (7)} & Qwen1.5B & \bfseries 0.60 & 57 & Qwen1.5B & 0.03 & \bfseries 50 & Qwen1.5B & 0.44 & 54 \\ \midrule

Winowhy & \makecell[l]{Qwen1.5B (7) \\ Llama1B (1)} & \makecell[l]{req (8) \\ instr (8)} & Qwen1.5B & \bfseries 0.62 & 71 & Qwen1.5B & 0.03 & \bfseries 61 & Llama1B & 0.44 & 62 \\ \bottomrule
\end{tabular}%
} 
\end{table*}

Analyzing the prompt component structure, as detailed in the `Dominant Components' section of Table \ref{tab:pareto_summary_detailed_restructured}, we observe that requests and instructions \textit{<req>, <instr>} were dominant across most tasks (Causal Judgment, Implicatures, Logical Fallacy, Navigate, Winowhy). They frequently yielded high-accuracy solutions and balanced knee points. This strongly suggests that providing clear task specifications and output formatting instructions is paramount for guiding SLMs effectively on these binary classification reasoning tasks.

The context component \textit{<context>} was key for achieving the lowest token counts in Hyperbaton and Logical Fallacy. It also featured contributing to the Causal Judgment front. While efficient, relying solely on context often resulted in lower accuracy compared to instruction-based prompts.

The Examples component \textit{<examples>} appeared necessary for top performance in Hyperbaton and Causal Judgment, indicating the value of few-shot learning for specific tasks, although it likely contributes to higher token counts.

Chain-of-Thought \textit{<cot>} hints were present in the highest-accuracy solution for Causal Judgment and efficient Hyperbaton solutions. This suggests that while not always necessary for these binary tasks with SLMs, prompting for step-by-step reasoning can be beneficial for more complex causal or grammatical judgments.

\section{Conclusion and future works} \label{Conclusion and future works}
We have applied our model and prompt grammar-guided evolutionary search using NSGA-II. The results quantitatively demonstrate the crucial trade-off between accuracy and token efficiency and reveal task-specific affinities for certain small language models and prompt structures. Although clear instructions (req, instr) proved broadly effective, components such as context, examples, and CoT offered advantages for specific tasks or efficiency goals. 

The generated Pareto fronts provided a practical tool, offering practitioners and decision-makers a range of optimized solutions. It helps them choose based on their specific performance requirements and cost constraints.

Building upon these contributions, several promising ideas for future research emerge. Firstly, extending the investigation beyond SLMs to include Large Language Models (LLMs) would be crucial for understanding how these techniques adapt to more powerful, but often more resource-intensive, models. Secondly, the exploration of alternative multi-objective evolutionary algorithms, such as MOEA/D or NSGA-III, could yield different trade-off characteristics.

Recognizing that, the selection of optimal prompts and models is inherently a multi-faceted challenge, future work should also broaden the scope of optimization objectives. Incorporating critical metrics such as fairness, bias, correctness, system latency, and others.

A detailed comparative analysis of our evolutionary approach against prominent prompt optimization techniques, benchmarking against gradient-based methods, reinforcement learning strategies, and LLM-as-judge pipelines. Distinguishing our method, and assessing its efficiency, cost, and performance are relevant.

 This automated approach eliminates the need for manual prompt, moving toward automatic prompt discovery and enabling practitioners to explore optimal prompt configurations rather than through trial and error. This work offers an initial foundation, outlining future directions for creating efficient AI systems capable of tackling increasingly complex tasks.



%
%
\bibliographystyle{splncs04}
\bibliography{bracis_biblio}
\end{document}